\documentclass[a4paper, 10 pt, conference]{ieeeconf}  
\IEEEoverridecommandlockouts                              
\usepackage{cite}
\usepackage{amsmath,amssymb,amsfonts}
\usepackage{algorithmic}
\usepackage{graphicx}
\usepackage{subcaption}
\usepackage{textcomp}
\usepackage{xcolor}
\usepackage{multirow}
\def\BibTeX{{\rm B\kern-.05em{\sc i\kern-.025em b}\kern-.08em
    T\kern-.1667em\lower.7ex\hbox{E}\kern-.125emX}}
\title{DeepBbox: Accelerating Precise Ground Truth Generation for Autonomous Driving Datasets\\
}

\author{Govind Rathore, Wan-Yi Lin and Ji Eun Kim
\thanks{Robert BOSCH LLC, 2555 Smallman Street, Pittsburgh, PA, 15222, USA
{\tt\small Govind.Rathore, Wan-Yi.Lin, JiEun.Kim@us.bosch.com}}}%

\begin{document}
\maketitle
\thispagestyle{empty}
\pagestyle{empty}

\begin{abstract}
Autonomous driving requires various computer vision algorithms, such as object detection and tracking. Precisely-labeled datasets (i.e., objects are fully contained in bounding boxes with only a few extra pixels) are preferred for training such algorithms, so that the algorithms can detect exact locations of the objects. However, it is very time-consuming and hence expensive to generate precise labels for image sequences at scale. In this paper, we propose DeepBbox, an algorithm that "corrects" loose object labels into right bounding boxes to reduce human annotation efforts. We use Cityscapes \cite{cordts2016cityscapes} dataset to show annotation efficiency and accuracy improvement using DeepBbox. Experimental results show that, with DeepBbox, we can increase the number of object edges that are labeled automatically (within 1\% error) by 50\% to reduce manual annotation time. 
\end{abstract}

\section{Introduction} \label{sec:intro}
\begin{figure*}
    \centering
    \includegraphics[width=0.5\textwidth]{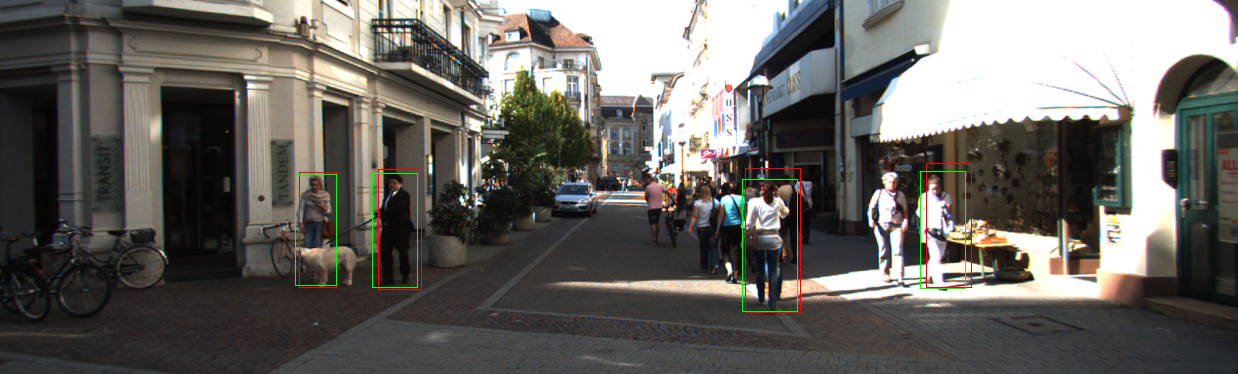}\includegraphics[width=0.5\textwidth]{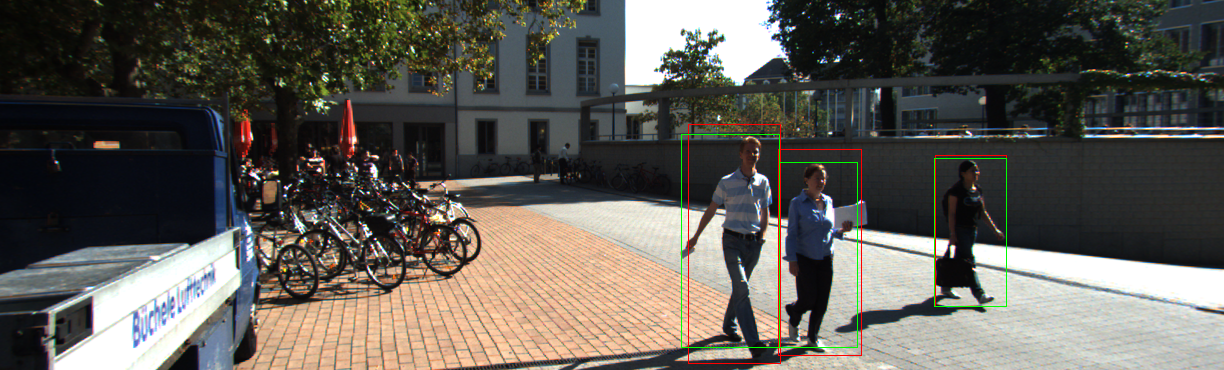}
    \caption{Example images from KITTI: labels provided by the dataset (red) and labels refined by DeepBbox (green)}
    \label{fig:KITTI_example}
\end{figure*}

Computer vision algorithms for autonomous driving in the wild require not only high accuracy, but also many other properties such as low memory, fast computation, and robustness against adversarial input, either natural (acquired under harsh weather) or man-made. To build such algorithms, large scale precisely-labeled (e.g., tight bounding boxes) datasets are required so that the algorithms can be trained under full supervision. One particular requirement of autonomous driving datasets is that labels need to be precise so that the error made by the whole system can be reduced to minimal as well as separated for effective validation. Although several benchmark datasets of autonomous driving such as KITTI \cite{KITTI}, Cityscapes \cite{cordts2016cityscapes}, and Berkeley Deep Drive (BDD) \cite{yu2018bdd100k} are available to the public, they are not labeled precisely, as red boxes in Fig. \ref{fig:KITTI_example}. 
It is time-consuming and labor-intensive to annotate a large dataset precisely. For example, ImageNet reported on the average 55 seconds for a high-quality/precise bounding box, including the time that the bounding boxes being reviewed and re-drawn (BDD, Cityscapes, and KITTI do not report time on precise bounding boxes). For a dataset as large as BDD (1.8 million bounding boxes), this translates to about 27.5 thousand hours of human effort. 

One way to reduce annotation effort is to have objects pre-labeled \cite{yu2018bdd100k} by computer vision algorithms such as object detectors. But the computer vision algorithms will only provide rough bounding boxes and still need human annotators to review, adjust or re-draw. Therefore, we need a method to make rough bounding boxes precise. Note that this paper addresses annotations using bounding boxes because it is the most popular way to label an object in current practice.

To improve the quality of pre-labels and ultimately reduce annotation cost, we propose DeepBbox, which takes an image patch as input, and estimates a tight bounding box around the main object in the image patch as output. The input image patch is a cropped region from the original image based on estimated object location by an object detector or tracker. We design a deep convolutional network architecture that can efficiently reduce errors of the computer vision pre-labels and be transferred to new datasets with small amount of training data. DeepBbox can also be used to make rough bounding box labels of existing datasets more precise, as green boxes in Fig. \ref{fig:KITTI_example}.

The contribution of this paper is as follows:
\begin{itemize}
    \item The proposed DeepBbox increases percentages of per-label edges that are precise and do not need human effort from 25.1\% to 37.7\% for a precision tolerance of 1\% i.e., the absolute error in predicted bounding box edge is within 1\% of the true bounding box's longest edge length.
    \item The proposed DeepBbox can be adopted early in the process of annotating new sets of data. The data size needed to re-train/fine-tune DeepBbox to a new dataset is as small as 7.6 thousand bounding boxes.
    \item The proposed DeepBbox can be applied to various video annotation pipelines to refine either bounding boxes roughly drawn by human or pre-labeled by computer vision algorithms. 
\end{itemize}

\section{Related Work}
\subsection{Time to draw a precise bounding box}\label{sec:related_work_annotation}
Bounding boxes are the most widely-used object annotation methods for applications of object detection, recognition, and tracking. Studies have shown that even with advances of user interfaces \cite{extreme_click}, median time required to draw a bounding box is still between 7 to 35 seconds \cite{PASCAL_VOC}, depending on image quality, box precision requirement and annotation task design. The cited numbers only include drawing and verifying a bounding box. If including the time of re-drawing bounding boxes that failed validation, the time will increase to 55 seconds.

There is always a trade-off between annotation quality and time, no matter which user interface is used. The proposed DeepBbox is to reduce the trade-off by improving annotation quality with deep networks that make loose bounding boxes precise without any additional manual annotation effort. 
%% To the best of our knowledge, there is no public reporting on time drawing precise, i.e., objects are fully contained in bounding boxes with less than 2 extra pixels, bounding boxes. %%WL: we may need to show how much more time it is
\subsection{Methods to reduce annotation effort}\label{sec:related_work_prelabel}
\subsubsection{Click supervision}
A recent work to reduce annotation effort is having annotators click on the center point of the object instead of draw a bounding box around it. Click supervision is particularly useful for video sequences because the scenes are highly correlated and the same object will get multiple clicks. Such repetition will help subsequent processing of the clicks to have a better estimate of the object. Studies have shown that click supervision, instead of bounding box supervision, can reduce annotation time by 18 times, but the object detector \cite{papadopoulos2017click_training}\cite{mettes2016click_training} and semantic segmentor \cite{bearman2016click_training_ss} trained under it achieves slightly worse results than bounding box annotations.

%Our proposed method aims to reduce annotation time without annotation quality decrease due to computer vision algorithms that trained on the annotated dataset. We provide annotations with the same quality as human draws a tight bounding box by using DeepBbox to automatically adjust rough bounding boxes.
%JK: I tried to rephrase the paragraph above to below. Please feel free to revert if you don't like. 
Our proposed method aims to reduce annotation time without sacrificing annotation quality. Automatically adjusting rough bounding boxes through DeepBbox achieves the equivalent quality in annotation as human workers draw tight bounding boxes carefully.  

\subsubsection{Semi-auto annotation}\label{sec:related_work_crowdai}
Another way of reducing annotation time is to have computer vision algorithms pre-label the bounding boxes and have the human workers adjust or re-draw them. The Berkeley Deep Drive \cite{yu2018bdd100k} dataset reported that by pre-labeling the bounding boxes with an object detector that was trained over half of the dataset, the annotation time is reduced by 60\%. Such time reduction comes mainly from cases that the pre-labels already correctly annotate the object without adjustment from human workers. Our work further reduces the annotation time by improving precision of the pre-labels so that a higher portion of the bounding boxes do not need adjustment at all.

\section{Proposed approach}
\begin{figure*}[t]
\centering
\begin{subfigure}{\textwidth}
        \includegraphics[width=\textwidth]{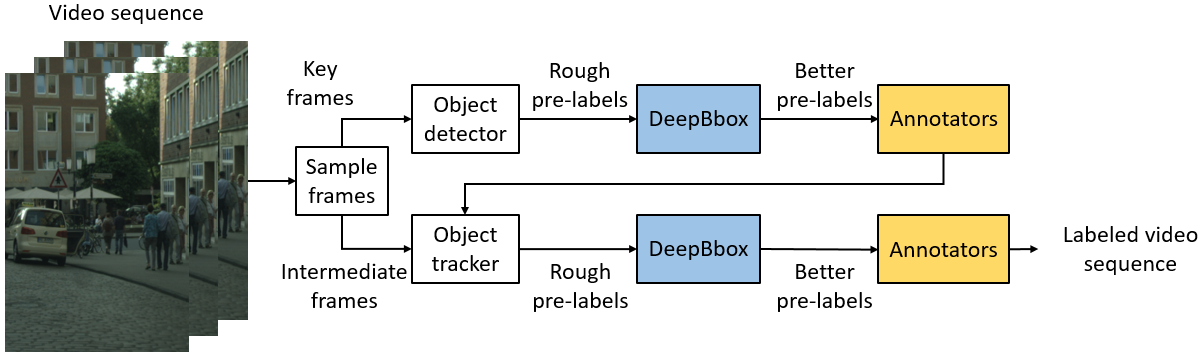}
        \caption{DeepBbox used in an annotation pipeline}
        \label{fig:pipeline_anno}
    \end{subfigure}
    \begin{subfigure}{\textwidth}
    \centering
        \includegraphics[width=0.9\textwidth]{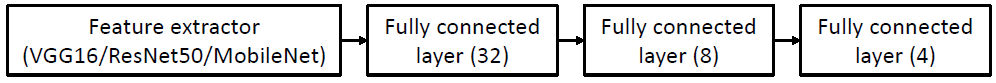}
        \caption{DeepBbox architecture}
        \label{fig:pipeline_arch}
    \end{subfigure}
\caption{Pipeline of the proposed method}
\label{fig:pipeline}
\end{figure*}

The use of the proposed DeepBbox for machine-in-the-loop data annotation is in Fig. \ref{fig:pipeline_anno}: input to DeepBbox (blue blocks) are image patches cropped from the video frames based on pre-labeled bounding boxes from computer vision algorithms, either object tracking or object detection. Before cropping the images, we expand the four edges of the pre-labeled bounding boxes to ensure that the visible part of the object is included in the image patch. DeepBbox then predicts precise bounding boxes for the objects. For illustration purposes, we use this common video annotation approach as an example, but DeepBbox can be applied to any annotation systems that utilizes computer vision pre-labels, or be used to make rough bounding boxes drawn by annotators precise.

 In such an annotation pipeline, input video sequence is first divided into key frames (sampled every $K$ frames, where $K$ can be determined by the speed of car movements and environmental changes) and intermediate frames. Pre-labels in key frames are initialized by object detectors then refined by DeepBbox. Pre-labels for key frames are reviewed then corrected or re-drawn by human to ensure key frame labels are precise. Annotated key frames are used to populate pre-labels for intermediate frames using object trackers. Pre-labels for intermediate frames are then refined by DeepBbox, and go through human annotators to correct them. 

\subsection{Architecture}\label{sec:arch}
High-level architecture of DeepBbox is shown in Fig. \ref{fig:pipeline_arch}: the first group of layers is the feature extractor, followed by three fully connected layers; the output dimension is four, which corresponds to the xy coordinates of top-left and bottom-right corners of the bounding box. Here, feature extractors refer to the convolutional layers of well-known deep neural network architectures for computer vision: VGG16 \cite{VGG16}, ResNet50 \cite{ResNet}, and MobileNet \cite{MobileNet}. 

Reasons for the above architecture are as follows:
\begin{itemize}
    \item Overall architecture: DeepBbox first learns/extracts features from image patches, and then uses such features to estimate the four extreme coordinates of the main object in the image patch. Hence we divide the architecture into a feature extractor and a coordinate estimator.
    \item Feature extractor: Design goal of the feature extractor is to represent the main object in the image patch. We start with experimenting on well-known feature extractors that have been shown effective in various computer vision tasks: the first 13 layers of VGG16, the first 49 layers of ResNet50, and the first 18 layers of MobileNet. Another selection metric of the feature extractor is transferability -- data annotation systems usually encounter images with different characteristics, therefore DeepBbox has to be easily transferable to different datasets. This can be viewed as requiring minimal training data for adaptation: as much as we make DeepBbox to be directly transferable to a new dataset, it will never perform as well as if it was trained on the same dataset. A common practice in data annotation, as what BDD \cite{yu2018bdd100k} has done, is to re-train or fine-tune computer vision algorithms used in the system after a portion of the data is annotated. One can see that the more the algorithms gets fine-tuned in the earlier stage during annotation, the more it can assist human annotators and hence lower the cost of annotating the rest of the dataset. Therefore, DeepBbox should require minimal data size to be fine-tuned to new datasets.
    
    Based on above considerations, in Section \ref{sec:exp_arch} we compare the three feature extractors with their trade-offs between estimation correctness and size of training data to seek an architecture with highest transferability.
    \item Coordinate estimator: Top layers of DeepBbox are to learn mappings between extracted image features and the four extreme coordinates of the main object in the image patch. To learn such nonlinear mapping, more than one layer of regression is needed. On the other hand we need to achieve better transferability, therefore we choose three layers with small number of nodes in each layer. 
    \item Loss function: The purpose of DeepBbox is to make bounding box boundaries as precise as possible, meaning that there are only a few pixels between the real object extreme points and the bounding box. Also, we intend to penalize small errors. Therefore, L1 distance (i.e., Manhattan difference) is the natural choice for measuring the performance of our estimation. To make optimization easier, we adopt Huber loss.
\end{itemize}
\begin{figure*}[t]
\centering
        \includegraphics[width=\textwidth]{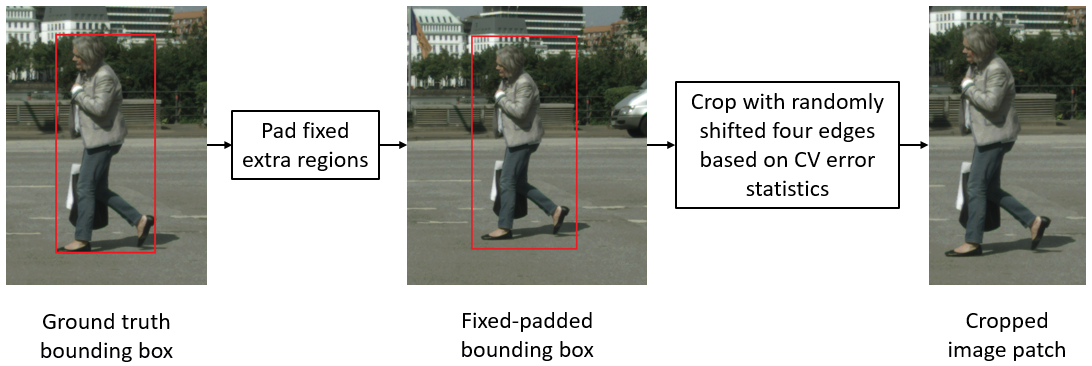}
\caption{Extracting image patches for training}
\label{fig:patch_extraction}
\end{figure*}
\subsection{Training procedure} \label{sec:training}
The first step of training DeepBbox is to estimate distributions of the error that DeepBbox is going to correct. Such error distribution helps the coordinate estimator to localize the main object in the image patch and the training procedure to mimic errors that DeepBbox needs to correct in the annotation pipeline. In addition, training with error distribution instead of real error from the computer vision algorithms better decouples DeepBbox from how exactly computer vision algorithms perform in the annotation pipeline. We will demonstrate these points in Section \ref{sec:exp_arch}.

Using the annotation pipeline in Fig. \ref{fig:pipeline_anno} as an example, the error that DeepBbox needs to correct will be the error present in predicted bounding boxes from object detector and tracker. Ideally, we should collect statistics of both algorithms, but to reduce training effort we can first train with the worse error. In this case, since the object tracker is initialized every $K$ frames and the objects of interest (vehicles and pedestrians) usually do not have sudden change of motion, as long as $K$ is not too large, bounding box edge error of the object tracker should be smaller than that of the object detector. Hence, we run the object detector over the whole training dataset and match bounding boxes with the ground truth to collect bounding box boundary error statistics introduced by the object detector.

\begin{figure*}\centering
\includegraphics[width=0.23\textwidth]{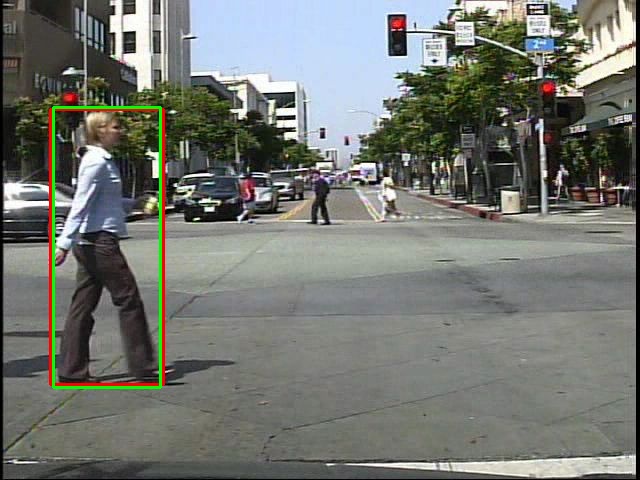}
\includegraphics[width=0.23\textwidth]{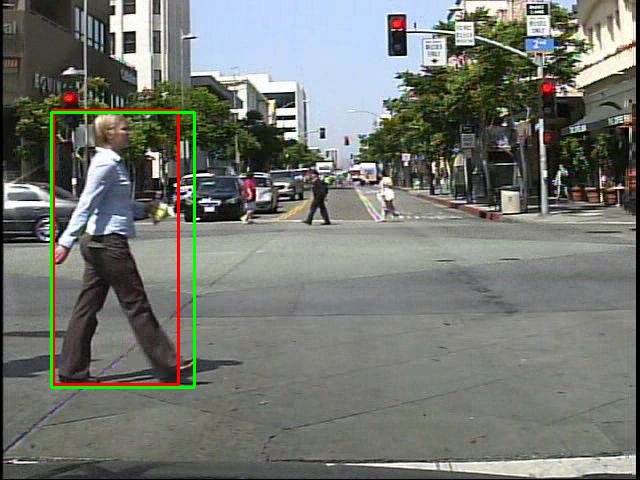}
\includegraphics[width=0.23\textwidth]{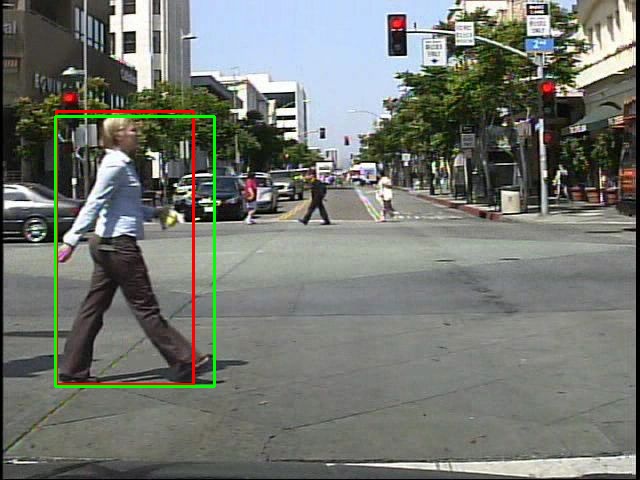}
\includegraphics[width=0.23\textwidth]{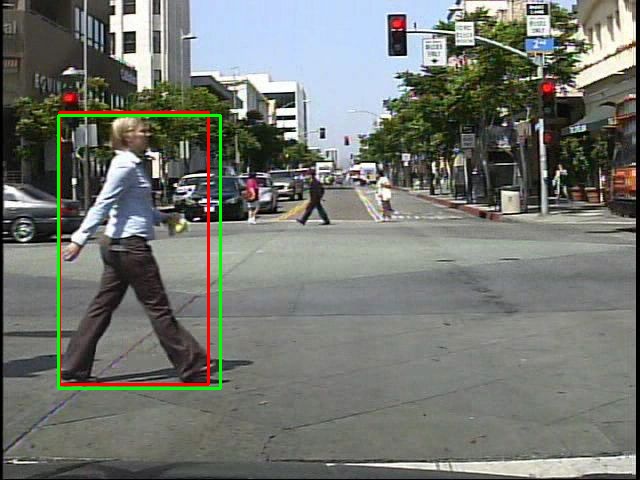}
\caption{DeepBbox correcting linear tracker on Caltech Pedestrian Detection dataset}\label{fig:tracker}
\end{figure*}

After obtaining computer vision error statistics, the next step is to extract image patches containing each object and ground truth coordinates of the object within each patch -- image patches will be the input of DeepBbox, and ground truth coordinates will be used to compute the loss value. The procedure of extracting image patches for training is illustrated in Fig. \ref{fig:patch_extraction}: given an image and ground truth bounding box of one fully-visible object, we first expand the four edges of the ground truth bounding box by a fixed ratio to ensure that the object is fully included in the image patch; then we shift the four edges (depending on the number drawn from the distribution, each edge can be moved inward or outward) randomly based on \textit{error statistics} collected from the object detector; we then crop the image patch, scale it to a common size and and normalize it by scaling the pixel values to be between [-1, 1]. The size scaling procedure maintains aspect ratio of the original patch and the extra region is padded with 0 values for all channels. Note that the training patches are generated on the fly, so the cropped patch size for a given true bounding box is different across training epochs.

\subsection{Known limitations and alternatives}
Since input image patches to DeepBbox is based on pre-labels by object detector and trackers with extra padding, one known limitation is occluded objects. In such scenario, the padded image patch may include more occluding objects than the objects being occluded, and the appearance of the occluded object may be very different from objects that are fully visible. If the occluding object is of the same class (ex: pedestrians occluding pedestrians), DeepBbox may mistake the occluding object as the main object and produce the bounding box for the occluding object; if the occluding object is of a different class (ex: cars occluding pedestrians), DeepBbox may make large error since appearance of the occluding object is unseen in the training set. Occluded objects are always hard for computer vision tasks, not only for DeepBbox, but also for object detector. In such cases, for annotation purposes, it is better not to use pre-labels and have human annotators draw the bounding box from scratch. 

Alternative methods to produce precise bounding boxes include:
\begin{itemize}
    \item Local image gradients: For pedestrian bounding boxes, using a pose-estimator we can locate the body parts that usually determine the edges of the bounding box; e.g. head and feet. Taking these key-points as seed, we could segment the pixels belonging to peripheral body parts through gradient based contour expansion \cite{active_contour}. For example, we can detect the pedestrian’s ankle and create a tiny circle around it as our initial contour. Iterative contour expansion would eventually engulf all the pixels belonging to pedestrian’s shoe. In the end, lowest point of the contour would give us the bottom edge of bounding box. Such method is very fast, requires a small set of example images (for tuning the expansion parameters) and hence can be transferred easily to new datasets. However, gradient based methods are difficult to generalize; e.g. the pedestrian hairline can be confused easily as an edge or the seed could be bad due to the input key-point being slightly outside the pedestrian contour. In our experiments, we found that this method does improve some bounding boxes but equally worsens others.
    \item Segmentation: Instead of learning the four extreme coordinates of the object, one can simply segment the input image patch, or the whole image frame, to obtain precise boundaries of the objects. However, segmentation is a much more difficult task than just estimation the four extreme coordinates because segmentation produces the full image mask. For annotating bounding boxes, segmentation will be an overkill. Also, difficulties of segmentation will result in much more complex models hence more training data is required resulting in lower performance when applied to new datasets.
\end{itemize}

\section{Experiments}
In this paper, we use pedestrians as our target object since they have more shape variations than vehicles hence more challenging to get precise bounding boxes. Using the example annotation pipeline in Fig. \ref{fig:pipeline_anno}, we show results on object detector pre-lables (key frames), and examples on object tracker pre-labels (intermediate frames). 

\subsection{Dataset and evaluation metrics}
Ideally, if we had access to pixel-precise bounding boxes then we can directly train an object detector on them. However, the publicly available object detection datasets do not yet have such high quality bounding boxes and therefore it is impossible to train a model that can produce tight bounding boxes. The Cityscapes \cite{cordts2016cityscapes} dataset has fine-grained semantic segmentation masks of objects but there are only 5000 such images (train, validation, test sets) in the dataset, presumably due to higher costs associated with finer annotations. An object detector trained entirely on such small dataset would likely not produce good results. In our work, we get around this limitation by ridding the object detector from the task of separating the True Positives from False Positives/Negatives. We use the Cityscapes fine-grained object masks from training set to generate tight bounding boxes for training our model that is meant to always predict a single bounding box.

The BDD \cite{yu2018bdd100k} dataset also contains fine-grained object segmentation masks for 8000 (training and validation sets) images. However, as of now, the dataset does not contain object instance segmentation. To extract the pedestrian bounding boxes from such segmentation masks, we first extract the boxes that had connected pedestrian pixels and then manually select the boxes that have a single fully-visible pedestrian. Due to limited time, we could only process about 25\% of BDD\textquotesingle s fine-grained images resulting in 652 tight bounding boxes. Due to the small number of bounding boxes from BDD, we only use it to demonstrate the transferability of DeepBbox.

We evaluate DeepBbox on two scenarios: the main test scenario is using DeepBbox to correct pre-labels from the well-known object detector Faster-RCNN \cite{faster_rcnn} to mimic the use of DeepBbox in annotation pipelines. We use object detector instead of object tracker because neither Cityscapes nor BDD has fine-grained segmentation masks for consecutive frames. We use Faster-RCNN pre-labels which have intersection over union $\geq$ $0.5$ between detected objects and ground through bounding boxes. Statistics of Cityscapes, BDD (the subset that we manually examined) and Faster-RCNN pre-labels can be found in Table \ref{table:Cityscapes_stats}. The secondary test scenario is expanding ground truth patches randomly to mimic different errors that DeepBbox needs to correct for different computer vision algorithms. 

From Faster-RCNN pre-lables on Cityscapes training dataset, we gather statistics of bounding box edge error for vertical and horizontal edges with respect to width and height respectively. Then approximate the vertical and horizontal edge error ratio as a Gaussian random variables: vertical edge error is $\mathcal{N}(0,0.0064)$ and horizontal edge error is $\mathcal{N}(0,0.0196)$. 

\begin{table}[]\centering
\begin{tabular}{|l|p{2cm}p{2cm}|}
\hline Dataset & Number of bounding boxes from dataset & Number of pre-labels from Faster-RCNN\\ \hline
Cityscapes train & 15219 &  6154\\ 
Cityscapes test & 3419 &  1363\\ 
BDD & 652 & 590\\\hline
\end{tabular}\caption{Dataset and Faster-RCNN pre-label statistics\label{table:Cityscapes_stats}}
\end{table}

To show how many bounding box edges were made precise and hence does not need human correction, we report percentages of bounding box edges within tolerance range before and after DeepBbox. In addition, we show the mean absolute error per edge in percentage of longest edge (denoted as MAE/LE) as an alternative evaluation metric.

\subsection{Results} \label{sec:exp_arch}
We compare performance of three feature extractors on Cityscapes. First, let's consider the scenario that training data and test data of DeepBbox are from the same dataset, and the error statistics (as described in Section \ref{sec:training}) used to train DeepBbox matches Faster-RCNN error. MAE/LE results are in the fourth column of Table \ref{table:Cityscapes_features}, and the percentage of precise bounding box edges is shown in Table \ref{table:Cityscapes_feature_tolerance}. Faster-RCNN MAE/LE error on Cityscapes is 4.25\% and on BDD is 3.11\%. Input image patches are scaled and padded to be 256x256 pixels, and test image patches are expanded according to the same distribution as training.

In Table \ref{table:Cityscapes_feature_tolerance} we show the performance with regard to tolerance error range from 1\% to 5\% of the longest edge. For example, the second cell of the second row of Table \ref{table:Cityscapes_feature_tolerance} means that 25.1\% of bounding box edges of Faster-RCNN pre-labels have an error within 1\% of the longest edge. 

From the fourth column of Table \ref{table:Cityscapes_features} and Table \ref{table:Cityscapes_feature_tolerance}, we can see that MobileNet and VGG16 achieve comparable correction performance (lowest number in Table \ref{table:Cityscapes_features} and highest number in Table \ref{table:Cityscapes_feature_tolerance}). According to Table \ref{table:Cityscapes_feature_tolerance}, DeepBbox is able to increase the percentage of bounding box edges within 1\% error range from 25.1\% to 37.7\%, which is about 50\% more. Such number means that compared to use Faster-RCNN pre-lables directly, 50\% more bounding box edges do not have to go through human correction.

\begin{table*}[]\centering
\begin{tabular}{|p{1.2cm}|p{2.5cm}p{2.5cm}|p{2.5cm}p{2.5cm}|}
\hline
Feature extractor  &        \multicolumn{2}{c}{Mismatch error statistics}  & \multicolumn{2}{|c|}{Matched error statistics} \\  \cline{2-5}
 &  Cityscapes trained with 30\% more error.& Cityscapes trained with 30\% less error  &  Cityscapes &BDD\\ \hline
VGG16  & 3.53     & 3.27 &\textbf{3.37} & 2.95 \\ 
MobileNet  &      3.32  & 3.32    &       \textbf{3.34} &3.04\\
ResNet50  &   3.70    &  3.25  &        3.62 & 3.16\\\hline
\end{tabular}\caption{Bounding box edge MAE/LE (\%) after DeepBbox correction on Cityscapes and BDD\label{table:Cityscapes_features}}
\end{table*}

\begin{table}[]
\begin{tabular}{|l|p{1.7cm}|p{1.4cm}p{1.5cm}p{1.4cm}|}
\hline Error & Faster-RCNN Pre-labels & DeepBbox (VGG16)  & DeepBbox (MobileNet) & DeepBbox (ResNet50) \\ \hline
1\% & 25.1 & 35.5 & 37.7 &33.5  \\
2\% &  42.6& 59.5 & 61.1 &  56.2\\
3\% &55.2  & 71.8 & 73.4 & 68.3 \\
4\% & 64.0 & 79.2 & 80.6 &  75.4\\
5\% & 71.4 & 84.2 & 85.1 & 80.9\\ \hline
\end{tabular}\caption{Percentage of bounding box edges within error tolerance before and after DeepBbox on Cityscapes test dataset\label{table:Cityscapes_feature_tolerance}}
\end{table}

Next we consider the case that DeepBbox was trained with different error in bounding box edges from Faster-RCNN as in the second and third column of Table \ref{table:Cityscapes_features}. These results are to show that same DeepBbox model can be used to correct different bounding box edge errors introduced by different computer vision algorithms (such as object tracker and detector), as shown in Fig. \ref{fig:pipeline_anno}. 

Although we are not able to run large-scale experiments for DeepBbox correcting error of object trackers, we show a few examples on Caltech Pedestrian Detection dataset \cite{caltech} in Fig. \ref{fig:tracker}. We manually annotate precise bounding boxes of the highlighted pedestrian in key-frames that are 5 frames apart, then pre-label the pedestrian for the in-between 4 frames via linear interpolation, shown as red boxes in Fig. \ref{fig:tracker}. The tracker pre-labels are then corrected by DeepBbox shown as green boxes in Fig. \ref{fig:tracker}.  Comparing the red boxes with green ones, we can see that DeepBbox corrected the top and bottom edges for all four frames, and corrected the right most edge for the right three frames.

\begin{figure}
    \centering
    \includegraphics[width=0.5\textwidth]{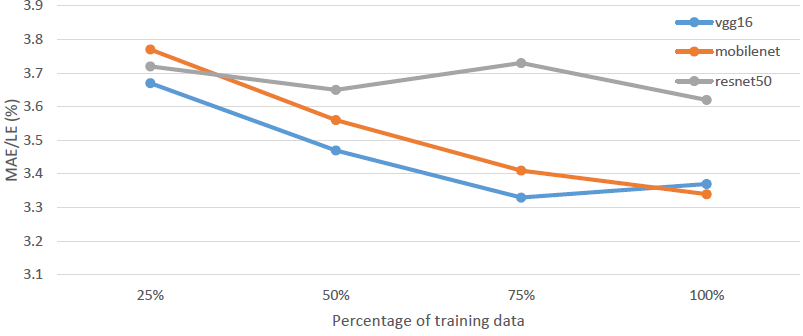}
    \caption{Performance of DeepBbox with partial Cityscapes training data}
    \label{fig:training_size}
\end{figure}

The last test case is that DeepBbox was trained on Cityscapes and test on BDD in the last column of Table \ref{table:Cityscapes_features}. As expected, without transfer learning, DeepBbox is at most improving MAE/LE by 10\% for Faster-RCNN pre-labels (MAE/LE is 3.11\%) on BDD, which means that at the very beginning of annotating new data, DeepBbox is not effective. Therefore, we run experiments on how much training data is needed to fine-tune DeepBbox to a new dataset in Figure \ref{fig:training_size}. We trained DeepBbox with 25\%, 50\%, 75\%, and full size of Cityscapes training set and test on Cityscapes. We can see that with VGG16 as feature extractor and 50\% of Cityscapes training set, it shows comparable results as utilizing full size of training set.  

%Finally, we show the performance of different hyper-parameters of DeepBbox. Table \ref{table:Cityscapes_inputsize} shows bounding box edge MAE percentage with respect to the longest edge for four different patch sizes. For Cityscapes image resolution 2048x1024, we found that normalizing the pedestrian image patches to 256x128 (height x width) yields the best performance with ResNet50 as feature extractor. However, it is clear that ResNet50 is more sensitive to input patch aspect ratio than the other two feature extractors. 

%\begin{table}[]\centering
%\begin{tabular}{|p{2cm}|p{1.5cm}p{1.5cm}p{1.5cm}|}
%\hline Patch dimension (height x width) &  DeepBbox (VGG16)  & DeepBbox (MobileNet) & DeepBbox (ResNet50) \\ \hline
%256x128 &  3.40 & 3.39 &\textbf{3.26}  \\
%256x256 &   3.37 & 3.34 &  3.62\\
%512x256 & 3.39 & 3.30 & 3.31 \\
%512x512 &  3.44 & 3.42 &  3.65\\\hline
%\end{tabular}\caption{DeepBbox performance (MAE/LE(\%)) with different patch size\label{table:Cityscapes_inputsize}}
%\end{table}

\section{Conclusions}
In this paper we present DeepBbox, a deep-learning-based algorithm that accelerates generation of precise ground truth for autonomous driving by making rough pre-labels tight around the object. We design the architecture that requires only half size of Cityscapes training set to train so it can be used early in the annotation pipeline. We increase the number of bounding box edges that do not need human annotation effort from 25.1\% to 37.7\%. We also show that DeepBbox can be used to correct error introduced by different computer vision algorithms used in data annotation.

\bibliographystyle{unsrt}
\bibliography{bib_deepbbox}

\end{document}